\def\year{2022}\relax
\documentclass[letterpaper]{article} 
\usepackage{aaai22}  
\usepackage{times}  
\usepackage{helvet}  
\usepackage{courier}  
\usepackage[hyphens]{url}  
\usepackage{graphicx} 
\usepackage{natbib}  
\usepackage{caption} 
\DeclareCaptionStyle{ruled}{labelfont=normalfont,labelsep=colon,strut=off} 
\usepackage{algorithm}
\usepackage{algorithmic}
\usepackage{amsmath}
\usepackage{amsfonts}
\usepackage{amssymb}
\usepackage{graphicx}
\usepackage{multirow}
\usepackage{setspace}

\usepackage[pagebackref=true,breaklinks=true,letterpaper=true,colorlinks,bookmarks=false,citecolor=cyan]{hyperref}
\usepackage{caption}
\usepackage{newfloat}
\usepackage{listings}
\floatstyle{ruled}
\newfloat{listing}{tb}{lst}{}
\floatname{listing}{Listing}

\title{Multi-Modal Perception Attention Network with Self-Supervised Learning for Audio-Visual Speaker Tracking}

\author{
    Yidi Li\textsuperscript{\rm 1}, 
    Hong Liu\textsuperscript{\rm 1}\thanks{Corresponding Author.},
    Hao Tang\textsuperscript{\rm 2}
}
\affiliations{
    \textsuperscript{\rm 1}Key Laboratory of Machine Perception, Peking University, Shenzhen Graduate School, China\\
    \textsuperscript{\rm 2}Computer Vision Lab, ETH Zurich, Switzerland\\
\{yidili, hongliu\}@pku.edu.cn, hao.tang@vision.ee.ethz.ch\\
}

\begin{document}

\maketitle

\begin{abstract}
Multi-modal fusion is proven to be an effective method to improve the accuracy and robustness of speaker tracking, especially in complex scenarios.
However, how to combine the heterogeneous information and exploit the complementarity of multi-modal signals remains a challenging issue.
In this paper, we propose a novel Multi-modal Perception Tracker (MPT) for speaker tracking using both audio and visual modalities.
Specifically, a novel acoustic map based on spatial-temporal Global Coherence Field (stGCF) is first constructed for heterogeneous signal fusion, which employs a camera model to map audio cues to the localization space consistent with the visual cues.
Then a multi-modal perception attention network is introduced to derive the perception weights
that measure the reliability and effectiveness of intermittent audio and video streams disturbed by noise.
Moreover, a unique cross-modal self-supervised learning method is presented to model the confidence of audio and visual observations by leveraging the complementarity and consistency between different modalities.
Experimental results show that the proposed MPT achieves 98.6\% and 78.3\% tracking accuracy on the standard and occluded datasets, respectively, which demonstrates its robustness under adverse conditions and outperforms the current state-of-the-art methods.
\end{abstract}

\section{Introduction}
Speaker tracking is the foundation task for intelligent systems to implement behavior analysis and human-computer interaction. 
To enhance the accuracy of the tracker, multi-modal sensors are utilized to capture richer information \cite{kilicc2017audio}.
Among them, auditory and visual sensors have received extensive attention from researchers as the main senses for human to understand the surrounding environment and interact with others.
Similar to the process of human multi-modal perception, the advantage of integrating auditory and visual information is that they can provide necessary supplementary cues \cite{xuan2020cross}.
Compared with the single-modal case, the utilizing of the complementarity of audio-visual signals contributes to improving tracking accuracy and robustness, particularly when dealing with complicated situations such as target occlusion, limited view of cameras, illumination changes, and room reverberation \cite{katsaggelos2015audiovisual}. Furthermore, multi-modal fusion shows distinct advantages when the information of one modality is missing, or neither modality is able to provide a reliable observation.
As a result, it is critical to develop a multi-modal tracking method that is capable of fusing heterogeneous signals and dealing with intermittent noisy audio-visual data.

\begin{figure}[t] \small
\centering
\includegraphics[width=1\columnwidth]{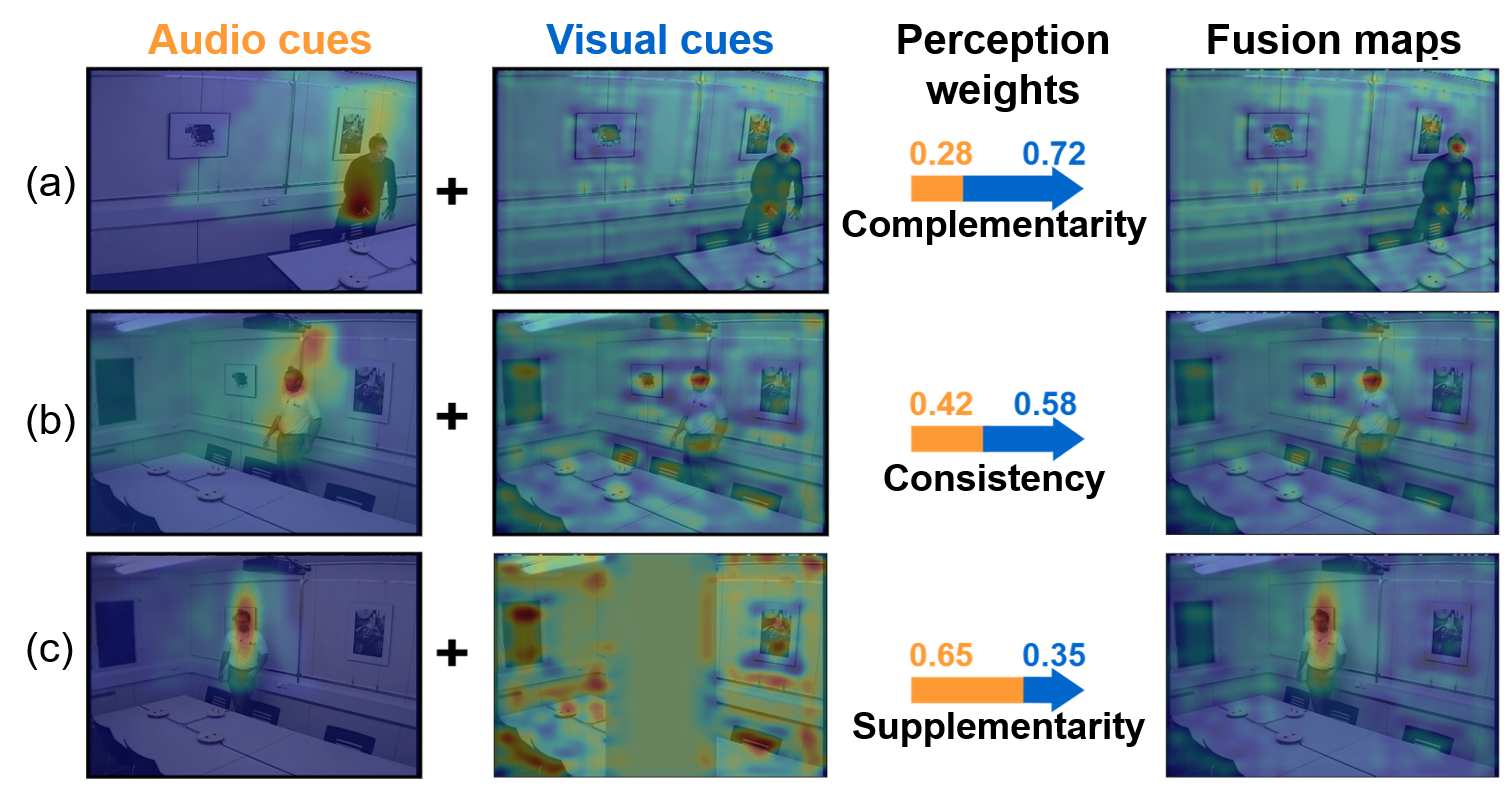} 
\caption{
Keyframes of the working process of the proposed multi-modal perception attention network. (a)-(c) demonstrate the exploration of complementarity, consistency, and supplementarity between audio-visual signals, respectively.}
\vspace{-0.4cm}
\label{fig:fig1}
\end{figure}

Current speaker tracking methods are generally based on probabilistic generation models due to their ability to process multi-modal information.
The representative method is Particle Filter (PF), which can recursively approximate the filtering distribution of tracking targets in nonlinear and non-Gaussian systems.
Based on PF implementation, the Direction of Arrival (DOA) angle of the audio source is projected onto the image plane to reshape the typical Gaussian noise distribution of particles and increase the weights of particles near DOA line \cite{kilicc2014audio}.
A two-layered PF is proposed to implement feature fusion and decision fusion of audio-visual sources through the hierarchical structure \cite{LYD2LPF}.
Moreover, a face detector is employed to geometrically estimate the 3D position of the target to assist in the calculation of the acoustic map \cite{qian2021audio}.
However, these methods prefer to use the detection results of the single modality to assist the other modality to obtain more accurate observations, while neglecting to fully utilize the complementarity and redundancy of audio-visual information.
In addition, most of the existing audio-visual trackers use generation algorithms \cite{ban2019variational,schymura2020audiovisual,qian20173d}, which are difficult to adapt to random and diverse changes of target appearance. 
Furthermore, the likelihood calculation based on the color histogram or Euclidean distance is susceptible to interference from observation noise, which limits the performance of the fusion likelihood.

To solve those limitations, we propose to adopt an attention mechanism to measure the confidence of multiple modalities, which determines the effectiveness of the fusion algorithm.
The proposed idea is inspired by the human brain's perception mechanism for multi-modal sensory information, which integrates the data and optimizes the decision-making through two key steps: estimating the reliability of various sources and weighting the evidences based on the reliability \cite{zhang2016decentralized}.
Take the intuitive experience as an example: when determining a speaker's position in a noisy and bright environment, we mainly use eyes; conversely, in a quiet and dim situation, we rely on sounds.
Based on this phenomenon, we propose a multi-modal perception attention network to simulate the human perception system that is capable of selectively capturing valuable event information from multiple modalities. 
Figure~\ref{fig:fig1} depicts the working process of the proposed network, in which the first two rows show the complementarity and consistency of audio and video modalities. In the third row, the image frame is obscured by an artificial mask to show the supplementary effect of the auditory modality when the visual modality is unreliable.
Different from existing end-to-end models, the specialized network focuses on perceiving the reliability of observations from different modalities.
However, the perception process is usually abstract, making it difficult to manually label quantitative tags.
Due to the natural correspondence between sound and vision, necessary supervision is provided for audio-visual learning \cite{hu2020discriminative} \cite{av-self}.
Therefore, we design a cross-modal self-supervised learning method, which exploits the complementarity and consistency of multi-modal data to generate weight labels of perception.

Neural networks have been widely used in multi-modal fusion tasks, represented by Audio-Visual Speech Recognition (AVSR) \cite{baltruvsaitis2018multimodal}.
However, except for preprocessing works such as target detection and feature extraction, neural network is rarely introduced to multi-modal tracking. 
This is because the positive samples in tracking task are simply random targets in the initial frame, resulting in a shortage of data to train a high-performing classifier.
Therefore, using an attention network specifically to train the middle perception component provides a completely new approach to this problem. 
Another reason is that the heterogeneity of audio and video data makes it difficult to accomplish unity in the early stage of the network. 
Therefore, we propose the spatial-temporal Global Coherence Field (stGCF) map, which maps the audio cues to the image feature space through the projection operator of a camera model.
To generate a fusion map, the integrated audio-visual cues are weighted by the perception weights estimated by the network.
Finally, a PF-based tracker improved with the fusion map is employed to ensure smooth tracking of multi-modal observations.

All these components make up our Multi-modal Perception Tracker (MPT), and experimental results demonstrate that the proposed MPT achieves significantly better results than the current state-of-the-art methods.

In summary, the contributions of this paper are as follows:
\begin{itemize}
\item A novel tracking architecture, termed Multi-modal Perception Tracker (MPT), is proposed for the challenging audio-visual speaker tracking task. Moreover, we propose a new multi-modal perception attention network for the first time to estimate the confidence and availability of observations from multi-modal data.
\item A novel acoustic map, termed stGCF map, is proposed, which utilizes a camera model to establish a mapping relationship between audio and visual localization space. Benefiting from the complementarity and consistency of audio-visual modalities, a new cross-modal self-supervised learning method is further introduced.
\item Experimental results on the standard and occluded datasets demonstrate the superiority and robustness of the
proposed methods, especially under noisy conditions.
\end{itemize}

\section{Related Works}
\subsubsection{Sound Source Localization.}
As the preprocessing module of many applications, Sound Source Localization (SSL) has been extensively studied.
Traditional microphone array-based acoustic sound source localization methods are based on Time Difference of Arrival (TDOA) \cite{cobos2020frequency}, steered beamforming \cite{chiariotti2019acoustic}, and high resolution spectral estimation \cite{yang2019multiple}.
Among these, the Generalized Cross-Correlation (GCC) algorithm is a commonly used TDOA estimation method, which describes the similarity between signals received at two sensors. Its reduced computational intensity leads to shorter decision time and higher tracking efficiency.
With the development of multi-modal technology, audio-visual learning is introduced to the SSL task.
Aiming at the problem of sound source localization in visual scenes, a two-stream network structure with attention mechanism is designed \cite{senocak2019learning}. The audio-visual category distribution matching method is developed to assist the selection and localization of the sounding object \cite{hu2020discriminative}.
Joint Deep Neural Networks (DNN) are proposed based on a probabilistic spatial audio model, including a visual DNN to localize candidate sound sources and an audio DNN to verify the localization of candidates \cite{masuyama2020self}.
We improve the Global Coherence Field (GCF) method to extract audio features with both spatial and temporal cues under the guidance of visual information.

\begin{figure*}[t] \small
\centering
\includegraphics[width=2\columnwidth]{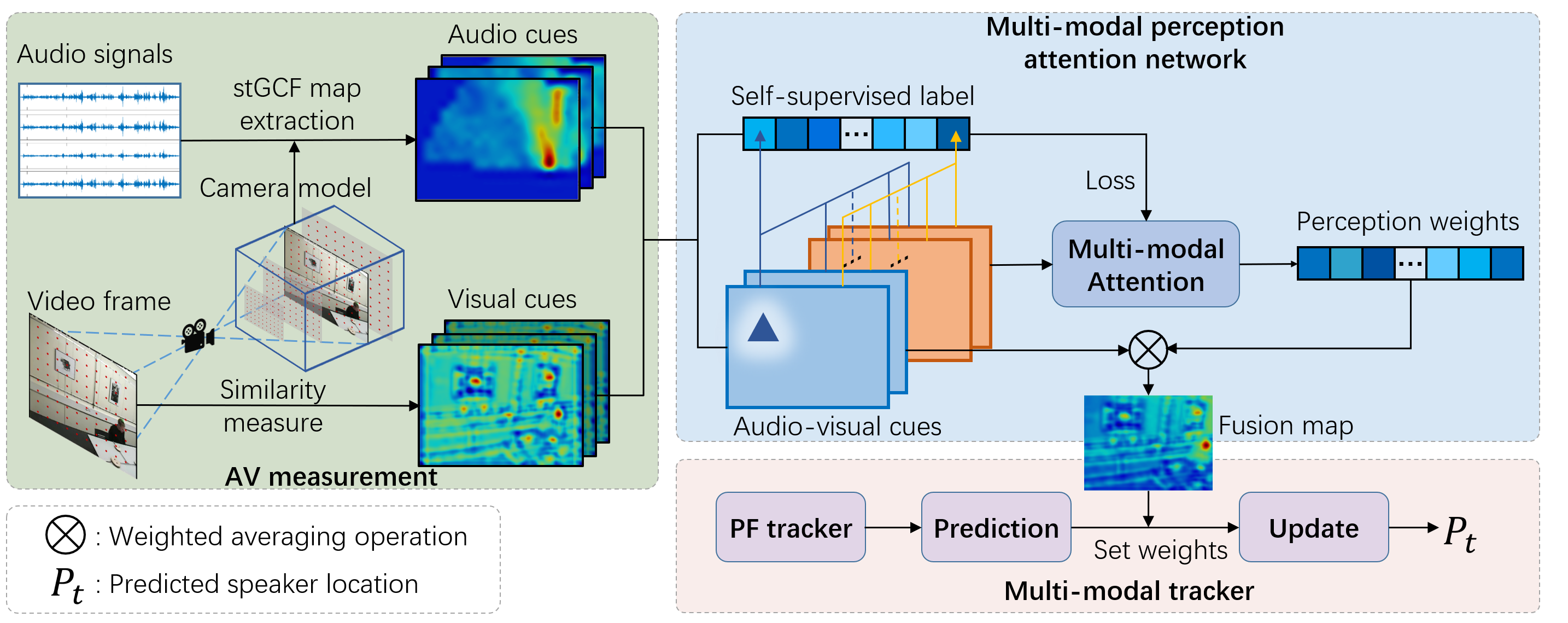} 
\caption{
The framework of the proposed tracking architecture. The stGCF-based audio cues are mapped to the localization space consistent with the visual cues. The integrated audio-visual cues combined with perception weights evaluated by the multi-modal perception attention network generate a fusion map that guides update step of the PF-based multi-modal tracker.
}
\vspace{-0.4cm}
\label{fig:fig2}
\end{figure*}

\subsubsection{Audio-Visual Tracking.}
Commonly used methods are state-space approaches based on the Bayesian framework.
Many works improve the PF architecture to integrate data streams from different modalities into a unified tracking framework. Among them, multi-modal observations are fused in a joint observation model, which is represented by improved likelihoods \cite{qian2019multi,kilicc2014audio,brutti2010joint}.
The tracking framework based on Extended Kalman Filter (EKF) realizes the fusion of an arbitrary number of multi-modal observations through dynamic weight flow \cite{schymura2020audiovisual}.
Probability Hypothesis Density (PHD) filter is introduced for tracking an unknown and variable number of speakers with the theory of Random Finite Sets (RFSs). The analytical solution is derived by introducing a Sequential Monte Carlo (SMC) implementation \cite{liu2019audio}.
By analyzing the task as a generative audio-visual association model formulated as a latent-variable temporal graphical model, a variational inference model is proposed to approximate the joint distribution \cite{ban2019variational}.
An end-to-end trained audio-visual object tracking network based on Single Shot Multibox Detector (SSD) is proposed, where visual and audio inputs are fused by an add merge layer \cite{wilson2020avot}.
Deep learning methods are less utilized in the audio-visual tracking task, leading to further research prospects.

\subsubsection{Attention-Based Models.}
Recently, the attention mechanism has been widely used in several tasks \cite{duan2021cascade,tang2021attentiongan,yang2021transformer,liu2021cross,duan2021audio,tang2019multi,xu2018structured}.
In visual object tracking, the Siamese network-based tracker is further developed by designing various attention mechanisms \cite{attention2018,attention2020}.
Based on the MDNet architecture, two modules of spatial attention and channel attention are employed to increase the discriminative properties of tracking \cite{attention2019}.
In audio-visual analysis, a cross-modal attention framework for exploring the potential hidden correlations of same-modal and cross-modal signals is proposed for audio-visual event localization \cite{xuan2020cross}.
For video emotion recognition, \cite{av-attention2} integrates spatial, channel and temporal attention into visual CNN, and temporal attention into audio CNN.
In audio-visual speech separation, the attention mechanism is used to help the model measure the differences and similarities between the visual representations of different speakers \cite{av-attention3}.
To the best of our knowledge, attention has not been studied on the audio-visual speaker tracking task.
In this paper, a self-supervised multi-modal perception attention network is introduced to investigate the perceptive ability of different modalities on the tracking scene.

\section{Proposed Method}
In this work, we propose a novel tracking architecture with a multi-modal perception attention network for audio-visual speaker tracking. Figure~\ref{fig:fig2} shows the overall framework of the proposed MPT, which consists of four main modules: audio-visual (AV) measurements, multi-modal perception attention network, cross-modal self-supervised learning, and PF-based multi-modal tracker.

\subsection{Audio-Visual Measurements}
Through audio-visual measurements, the corresponding cues are extracted from audio signals and video frames. To integrate multi-modal cues in the same state space, we map the audio cues to the same localization plane as visual cues.

\subsubsection{Audio Measurement.}
The acoustic map that highlights the source positions can be accomplished through a coherence measure based on cross-power spectrum phase, such as the Generalized Cross Correlation with Phase Transfrom (GCC-PHAT) \cite{gccphat}. On this basis, we introduce the stGCF method to extract the audio cues. 
Define $\mathbf{r}_{ik}^{PHAT}(t,\tau)$ as the GCC-PHAT derived from microphone pair $(i, k)$. It shows a prominent peak where the delay $\tau$ is equal to the actual TDOA. Let $\tau_{ik}(p)$ denote the theoretical time delay of a generic point $p$ relative to the microphone pair $(i,k)$. For the set $\Omega$ of $M$ microphone pairs, the GCF value is defined as the average of the GCC-PHAT for each microphone pair belonging to $\Omega$:
\begin{equation}
\mathbf{r}_{\Omega}^{GCF}(t,p)=\frac{1}{M}\sum\limits_{(i,k)\in\Omega}\mathbf{r}_{ik}^{PHAT}(t,\tau_{ik}(p)).
\end{equation}

Given a spatial grid with potential sound source positions, the GCF value represents the probability of the existence of a sound source at each position.
To construct the spatial grid, a pinhole camera model is utilized to project the 2D points on the image plane into a series of 3D points with different depths in 3D world coordinates, where the depth refers to the vertical distance from the 3D point to the camera's optical center.
Assuming that a set $D$ with $d$ depths is represented as $D=\{D_{k}, k = 1,...,d\}$, given depth $D_k$, the image-to-3D projection process is formulated as:
\begin{equation}
p^{3d}_{ijk} = \Phi(p^{2d}_{ij};D_{k}),
\label{eq:2}
\end{equation}
where $\Phi$ is the projection operator, $i$ and $j$ are the index of vertical and horizontal coordinate of the point, $i = 1,...,h$ and $j = 1,...,w$. A 2D sampling point set, $\mathbf{P}^{2d}=\{p^{2d}_{11},...,p^{2d}_{hw}\}$, is constructed by sampling on the image plane. Through Eq.~\eqref{eq:2}, $\mathbf{P}^{2d}$ is projected to multiple planes with different depths, $\mathbf{P}^{2d}\stackrel{\Phi}{\longrightarrow}\{\mathbf{P}^{3d}_{k},k =1,...,d\}$, where $\mathbf{P}^{3d}_{k}$ is the sample set on the plane with the depth $D_{k}$. The GCF map derived from $\mathbf{P}^{3d}_{k}$ is formulated as:
\begin{equation}
\mathbf{R}_{\Omega}^{GCF}(t,\mathbf{P}^{3d}_{k}) = \begin{bmatrix}
\mathbf{r}(p_{11k}) & \dots   & \mathbf{r}(p_{1wk})\\
   \vdots                    & \ddots & \vdots                       \\
\mathbf{r}(p_{h1k}) & \dots    & \mathbf{r}(p_{hwk}) 
\end{bmatrix}_{h \times w},
\end{equation}
where $\mathbf{r}(p_{\cdot \cdot k})$ is short for $\mathbf{r}_{\Omega}^{GCF}(t,p_{\cdot \cdot k}^{3d})$. Assuming that the peak of GCF map is at the $k_{max}$-th depth, the spatial GCF (sGCF) map at time $t$ is defined as:
\begin{equation}
\mathbf{R}_{\Omega}^{sGCF}(t,\mathbf{P}^{3d}) = \mathbf{R}_{\Omega}^{GCF}(t,\mathbf{P}^{3d}_{k_{max}}). 
\end{equation}
Due to the intermittent nature of speech and the continuity of the speaker's movement, the speech signals over a period provide references for audio cues at the current moment. Considering the signal in the time interval $[t-m_{1},t]$, the $m_{2}$ frames with largest peak values of sGCF maps are selected among $m_{1}+1$ frames.
The stGCF map at time $t$ is defined as:
\begin{equation}
\mathbf{R}_{\Omega}^{stGCF}(t,\mathbf{P}^{3d}) = \{\mathbf{R}_{\Omega}^{sGCF}(t',\mathbf{P}^{3d})|t'\in \mathbf{T}\}, 
\end{equation}
where $\mathbf{T}$ denotes the time set of the $m_{2}$ frames.

\subsubsection{Visual Measurement.}
The tracking task aims to localize an arbitrary target selected in the first frame of the video, which makes it impossible to collect data in advance to train a specific detector for tracking.
 Therefore, the general deep metric learning method is introduced to train the model at the initial offline stage, which considers the tracking problem as the similarity measurement between a known target and the search area. A pre-trained Siamese network \cite{simFC} is employed in this module, which uses cross-correlation as the metric function completed by the convolution operation.
The output response maps are equipped as visual cues, which can be formulated as:
\begin{equation}
\mathbf{S}(I_{t}) = \{f(I_{t},I^{ref})|I^{ref}\in \mathbf{I}\},
\end{equation}
where $I_{t}$ is the current video frame, $I^{ref}$ is the reference template which is the user-defined tracking target in the first frame, and $\mathbf{I}$ is the set of the reference templates with different scales. $f(\cdot)$ denotes the metric function that outputs a representative score map.
The $\mathbf{S}(I_{t})$ reflects the probability of the tracking target at each position in the search image, which is consistent with the meaning of the stGCF maps referring to the audio cues.

\subsection{Multi-Modal Perception Attention Network}
Given the extracted audio and visual cues, the multi-modal perception attention network (see Figure~\ref{fig:fig2}) generates a confidence score map as a speaker location representation.
The brain's attention mechanism is able to selectively improve the transmission of information that attracts human attention, weighing the specific information that is more critical to the current task goal from abundant information.
Inspired by this signal processing mechanism, a neural attention mechanism is exploited in this module to learn to measure the plausibility of multiple modalities.

To integrate the audio and visual cues, the stGCF maps $\mathbf{R}_{\Omega}^{stGCF}$ and visual response maps $\mathbf{S}(I_{t})$ are normalized and reshaped into 3D matrix form, expressed as:
\begin{equation}
\begin{split}
&\mathbf{R}=[\mathbf{R}_{1},...,\mathbf{R}_{D^{a}}]\in\mathbb{R}^{U\times D^{a}}, \\
&\mathbf{S}=[\mathbf{S}_{1},...,\mathbf{S}_{D^{v}}]\in\mathbb{R}^{U\times D^{v}},
\end{split}
\end{equation}
where $U$ denotes the size of each input video frame, $U=H\times W$. $D^{a}$ is the dimension of the audio cues, which depends on $m_{2}$ referring to temporal cues, and $D^{v}$ is dimension of the video cues, which is determined by the number of $I^{ref}$.
The fused audio-visual cues, $\mathbf{V} = [\mathbf{R}_{1},...,\mathbf{R}_{D^{a}},\mathbf{S}_{1},...,\mathbf{S}_{D^{v}}]$, are processed through a base network, which draws on the architecture of the channel attention module \cite{CBAM}, where the channel corresponds to the observation extracted from the audio or visual modality. For each channel $i\in\{1,...,D^{a}+D^{v}\}$, the attention mechanism $G_{att}$ generates a positive score $\alpha_{i}$ to measure the reliability of the observation on the $i$-th channel. The processing is formulated as:
\begin{equation}
G_{att}(\mathbf{V})=[\alpha_{1},...,\alpha_{D^{a}+D^{v}}]\in\mathbb{R}^{1\times (D^{a}+D^{v})},
\end{equation}
where the score $\alpha_{i}$, termed the perceptual weight, reflects the confidence level of the multi-modal cues measured according to the previous section. The $\alpha_{i}$ is higher in reliable observations and turns to lower in ambiguous observations disturbed by background noise, room reverberation, visual occlusion, confusing background, etc. This gets benefits from the statistical features learned by the network from observation maps. Through this, the network exhibits the perceived ability to multi-modal observations, which describes the working interpretability of the proposed network.

\subsection{Cross-Modal Self-Supervised Learning}
The sensing capability accomplished by the network is an abstract process, which makes it impossible to label data artificially for essential supervision. To this end, a new cross-modal self-supervised learning strategy is proposed to train the network.
The self-supervision includes a temporal factor and a spatial factor, which consider the temporal continuity of moving targets and the positional consistency in multi-modal observations, respectively.
For the $i$-th channel, assuming that point $p_{t,i}^{max}$ is the position of the peak of the feature map at time $t$, the corresponding spatial factor of the observation on channel $i$ is defined as the averaging operators in and across the multiple modalities. The cross-modal spatial factor is formulated as:
\begin{equation}
l_{t,i}^{s}=\frac{1}{2}[\frac{1}{D^a}\sum\limits_{j=1}^{D^{a}}\mathbf{R}_{t,j}(p_{t,i}^{max})+\frac{1}{D^v}\sum\limits_{k=1}^{D^{v}}\mathbf{S}_{t,k}(p_{t,i}^{max})],
\end{equation}
where $\mathbf{S}_{t,k}(p)$ denotes the normalized visual response value at position $p$.
Note that $\mathbf{R}_{t,j}(p)$ is the normalized sGCF value at position $p$, where $j$ is the depth index. 

\begin{figure}[t] 
\centering
\includegraphics[width=1\columnwidth]{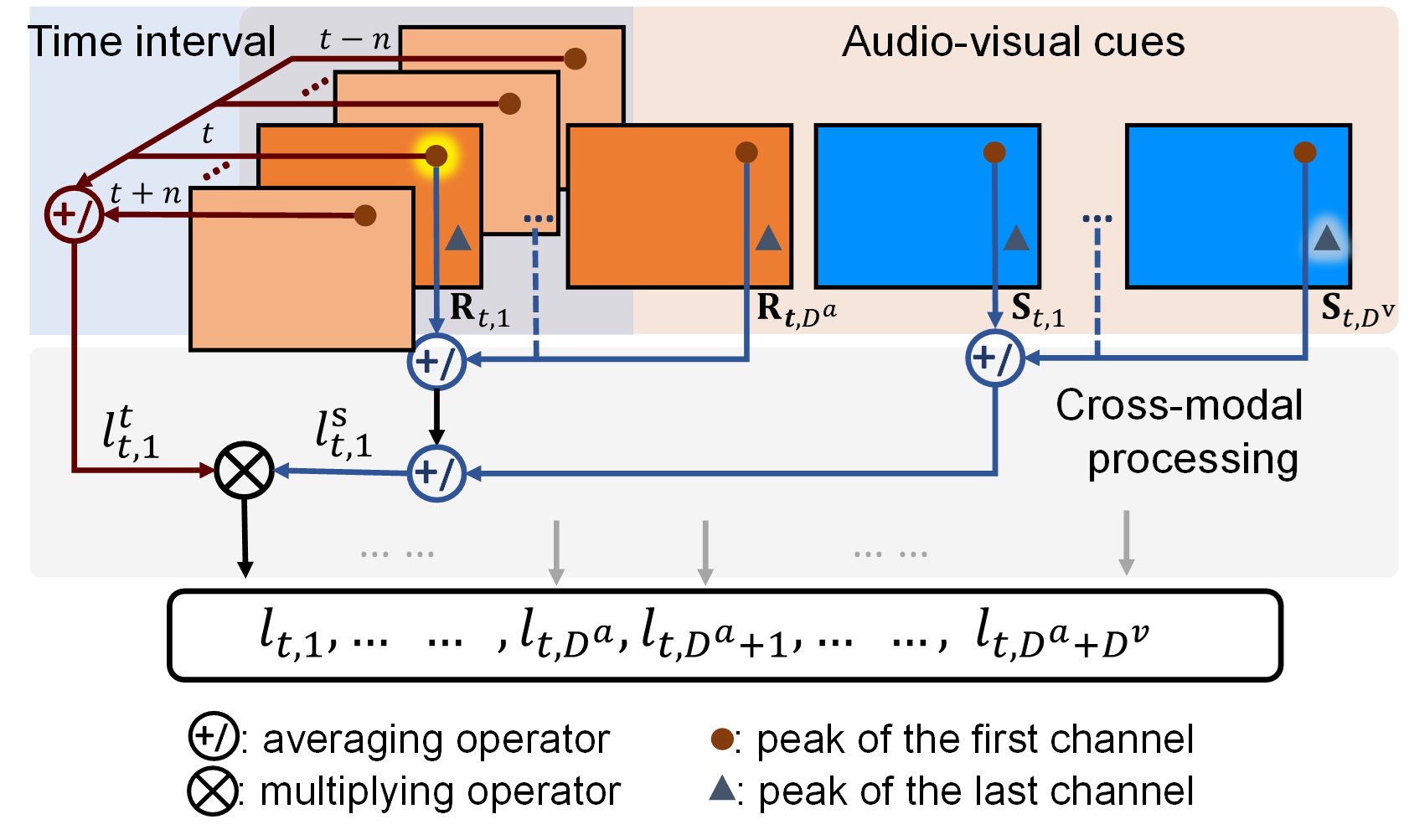} 
\caption{
Illustration of the self-supervised labels generated across modalities in a time interval.
}
\vspace{-0.4cm}
\label{fig:3}
\end{figure}

The temporal factor is derived by performing the above averaging operation on a time interval centered on time $t$. The temporal factor and the self-supervised label are expressed as:
\begin{equation}
\begin{split}
& l_{t,i}^{t}=\frac{1}{2n+1}\sum\limits_{q=t-n}^{t+n}\mathbf{V}_{q,i}(p_{t,i}^{max}),\\
& l_{t,i}=l_{t,i}^{s}\times l_{t,i}^{t},
\end{split}
\end{equation}
where $\mathbf{V}$ denotes the audio map or visual map. 
As shown in Figure \ref{fig:3}, the self-supervised label integrates the evaluations from different modalities in a time interval.
When the target drifts on one observation, according to the complementarity between the modalities and the continuity of target motion, the lower value is provided by the other channel with more accurate observation. In addition, when the peaks of all observations are located in the same area, the value will increase accordingly.
The general L2 loss is used to evaluate the generated labels and the attention measures.
\subsection{Multi-Modal Tracker}
The attention network introduced above supports multi-modal tracking through an improved PF algorithm. 
The attention measure output by the network is used to weight the audio-visual cues $\mathbf{V}$.
Compared with the traditional additive likelihood and multiplicative likelihood, the weighting method based on the attention mechanism is essentially closer to the human sensory selective attention mechanism.
Fusion map obtained after weighted average is expressed as:
\begin{equation}
\mathbf{Z}=\frac{1}{D^{a}+D^{v}}\sum\limits_{i=1}^{D^{a}+D^{v}}\alpha_{i}\mathbf{V}_{i}.
\end{equation}
The perception attention values of different modalities are fused in the map and used to weight particles in the update step of the PF. After diffusion, the value of the fusion map at the particle position is set as the new particle weight.
Moreover, in order to utilize the global information of the fusion map, we simply improve the resampling step as well. At the beginning of each iteration, a group of the particles is reset to the peak position of the fusion map. Through the correction of the peak value, the tracking drift problem caused by the observation noise of some frames is avoided. The method is outstanding when the observation is severely disturbed by the environment noise.

\begin{table*}[]
\centering
\renewcommand\arraystretch{1.2}
\small 
\tabcolsep1.9mm
\begin{tabular}{|c|c|cc|c|c|c|c|c|c|c|c|c|c|c|}
\hline
\multicolumn{2}{|c|}{Sequences}& \multicolumn{2}{c|}{Uni-modal}& \multicolumn{4}{c|}{Multi-modal}& \multicolumn{2}{c|}{Uni-modal+Occ}                                        & \multicolumn{4}{c|}{Multi-modal+Occ}                                                                                                                  & \multirow{3}{*}{\begin{tabular}[c]{@{}c@{}}Occ \\ rate\end{tabular}} \\ \cline{1-14}
\multirow{2}{*}{Seq} & \multirow{2}{*}{Cam} & \multicolumn{1}{c|}{AO}     & VO    & AV-A  & AV3D  & 2LPF & MPT(ours)       & \multicolumn{2}{c|}{VO}                                                   & \multicolumn{2}{c|}{2LPF}                                                 & \multicolumn{2}{c|}{MPT(ours)}                                                  &                                                                      \\ \cline{3-14}
                     &                      & \multicolumn{2}{c}{MAE$ \downarrow$} & \multicolumn{4}{|c|}{MAE$ \downarrow$} & \multicolumn{1}{l|}{MAE$ \downarrow$} & \multicolumn{1}{l|}{ACC$ \uparrow$} & \multicolumn{1}{l|}{MAE$ \downarrow$} & \multicolumn{1}{l|}{ACC$ \uparrow$} & \multicolumn{1}{l|}{MAE$ \downarrow$} & \multicolumn{1}{l|}{ACC$ \uparrow$} &                                                                      \\ \hline
                     & 1                    & \multicolumn{1}{c|}{32.87}  & 21.41 & 10.75 & 4.31  & 3.32 & 3.67          & 103.46                               & 26.35                              & 94.45                                & 42.97                              & 11.54                                & 88.79                              & 29.88                     \\ \cline{2-2}
08                   & 2                    & \multicolumn{1}{c|}{18.76}  & 16.58 & 7.33  & 4.66  & 3.06 & 3.58          & 181.82                               & 19.14                              & 75.41                                & 62.42                              & 7.88                                 & 92.75                              & 37.74                                                                \\ \cline{2-2} 
                     & 3                    & \multicolumn{1}{c|}{27.01}  & 15.73 & 9.85  & 5.34  & 3.47 & 3.43          & 141.76                               & 21.53                              & 68.54                                & 50.51                              & 14.3                                 & 81.52                              & 54.62                      \\ \hline
                     & 1                    & \multicolumn{1}{c|}{28.27}  & 14.69 & 14.66 & 8.15  & 6.15 & 6.77          & 30.12                                & 81.06                              & 26.35                                & 82.91                              & 13.57                                & 88.93                              & 15.66                      \\ \cline{2-2} 
11                   & 2                    & \multicolumn{1}{c|}{24.16}  & 16.42 & 14.01 & 7.48  & 5.58 & 4.55          & 116.87                               & 19.45                              & 111.47                               & 27.31                              & 26.06                                & 65.65                              & 70.17                                                                \\ \cline{2-2} 
                     & 3                    & \multicolumn{1}{c|}{25.66}  & 21.54 & 13.96 & 6.64  & 3.86 & 3.84          & 86.60                                & 42.95                              & 49.97                                & 50.00                              & 21.98                                & 77.40                              & 66.32                      \\ \hline
                     & 1                    & \multicolumn{1}{c|}{40.67}  & 17.83 & 12.49 & 6.86  & 4.11 & 4.67          & 93.07                                & 39.88                              & 122.72                               & 16.87                              & 17.43                                & 77.71                              & 32.55                      \\ \cline{2-2} 
12                   & 2                    & \multicolumn{1}{c|}{24.26}  & 19.03 & 10.81 & 10.67 & 5.39 & 4.84          & 145.54                               & 23.12                              & 104.3                                & 31.15                              & 23.96                                & 65.75                              & 74.50                                                                \\ \cline{2-2} 
                     & 3                    & \multicolumn{1}{c|}{34.02}  & 22.29 & 11.86 & 9.71  & 5.65 & 3.78          & 157.37                               & 21.78                              & 144.25                               & 25.48                              & 21.97                                & 66.57                              & 78.35                      \\ \hline
\multicolumn{2}{|c|}{Average}               & \multicolumn{1}{c|}{28.40}  & 18.39 & 11.74 & 7.091 & 4.51 & \textbf{4.34} & 117.40                               & 32.80                              & 88.60                                & 43.29                              & \textbf{17.63}                       & \textbf{78.34}                     & 51.08                                                                \\ \hline
\end{tabular}	
\setlength{\abovecaptionskip}{0.1cm} 
\caption{Comparison results with uni-modal methods and the state-of-the-art audio-visual methods on the original dataset and the occluded dataset. Occ rate is the percentage of frames in which the speaker is occluded by the mask. MAE is in pixel, ACC is in \%. The proposed method achieves robust tracking in the presence of occlusion. (Occ: occluded sequences, AO: audio-only, VO: visual-only, AV-A: \cite{kilicc2014audio}, AV3D: \cite{qian20173d}, 2LPF: \cite{LYD2LPF}, MPT: ours )
	}  
\vspace{-0.4cm}
\label{table:1}
\end{table*}

\begin{table}[]
\centering
\setlength{\abovecaptionskip}{0.1cm} 
\renewcommand\arraystretch{1.2}
\small 
\tabcolsep1.8mm
\begin{tabular}{|c|c|c|c|c|c|c|}
\hline
\multirow{2}{*}{AM} & \multirow{2}{*}{AN} & \multirow{2}{*}{TR} & \multicolumn{2}{c|}{Org}        & \multicolumn{2}{c|}{Occ}        \\ \cline{4-7} 
                    &                     &                     & MAE$\downarrow$           & ACC$\uparrow$            & MAE$\downarrow$           & ACC$\uparrow$           \\ \hline
GCF                 & -                   & -                   & 80.15          & 45.23          & 80.15          & 45.23          \\ \hline
stGCF               & -                   & -                   & 28.40          & 63.58          & 28.40           & 63.58          \\ \hline
stGCF               & AvgAtt              & -                   & 22.63          & 74.16          & 33.48          & 60.59          \\ \hline
stGCF               & MPAtt               & -                   & 12.56          & 89.50           & 24.57          & 72.17          \\ \hline
stGCF               & AvgAtt               & IPF                 & 17.33         & 78.22          & 26.88          & 67.54          \\ \hline
\textbf{stGCF}               & \textbf{MPAtt}               & \textbf{IPF}                 & \textbf{4.34} & \textbf{98.55} & \textbf{17.63} & \textbf{78.34} \\ \hline
\end{tabular}

	\caption{Influence of each innovative component in MPT, compared with the general GCF feature and average attention. (AM: audio measurement, AN: attention network, TR: tracker, AvgAtt: average attention, MPAtt: multi-modal perception attention, IPF: improved PF, Org: original dataset)
	}  
\vspace{-0.4cm}
\label{table:2}
\end{table}

\section{Experiments and Discussions}
\subsubsection{Datasets.}
In this section, the proposed tracker is evaluated on the AV16.3 corpus \cite{av163}, which provides true 3D mouth location derived from calibrated cameras and 2D measurements on the various images for systematic assessment. The audio data is recorded at the sampling rate of 16 kHz by two circular eight-element microphone arrays placed 0.8m apart on the table. The images are captured by 3 monocular color cameras installed in 3 corners of the room at 25Hz with size $H\times W=288\times360$. The experiments are tested on $seq$08, 11, and 12, where a single participant wandered around, moved quickly, and spoke intermittently. Each set of experiments uses signals from two microphone arrays and an individual camera. 
\subsubsection{Implementation Details.}
Visual cues are generated by a pretrained Siamese network \cite{simFC} based on AlexNet backbone. Reference image set $\mathbf{I}$ contains two target rectangles with scales of 1 and 1.25, which are defined by users in the first frame.
For audio measurement, the number of 2D sampling points in the horizontal and vertical directions on the image plane are $w=20$ and $h=16$. A $0.8m$ high table is placed in a $(3.6\times8.2\times2.4)m$ room. Therefore, the sampling points located outside the room range and below the desktop are removed, which is in accord with the real situation and avoids the ambiguity caused by the symmetry of the circular microphone. The depths number of projected 3D points is set to $d=6$. The speech signal is enframed to $40ms$ by a Hamming window with a frame shift of $1/2$. The parameters to calculate stGCF are set to $M=120$, $m_{1}=15$, $m_{2}=5$.
Backbone of the attention network is MobileNetv3-large \cite{mobilenetv3}.
The network is trained on single speaker sequences $seq$01, 02, 03, which contain more than 4500 samples. The parameters to generate self-supervised label are set to $D^{a}=5$, $D^{v}=2$, $n=6$. All models are trained for 20 epochs with batch size 16 and learning rate 0.01.
Our method and comparison methods are based on Sampling Importance Resampling (SIR)-PF for tracking.
The number of particles is set to 100. Our source codes are available at
https://github.com/liyidi/MPT.

\subsubsection{Evaluation Metrics.}
Mean Absolute Error (MAE) and the Accuracy (ACC) is used to evaluate performance of tracking methods. MAE calculates the Euclidean distance in pixel between the estimated position and the ground truth (GT), divided by the number of frames. ACC measures the percentage of correct estimates, whose error distance in pixel does not exceed 1/2 of the diagonal of the bounding-box of GT.

\subsubsection{Comparison Results.}
The proposed MPT is compared with the uni-modal method and the state-of-the-art audio-visual methods, which are based on the PF architecture.
The AO and VO methods are implemented based on the audio cues and visual cues proposed in the previous section. 
Furthermore, in order to verify the robustness of the tracker under interference conditions, we conducted comparative experiments on the occluded data. The occlusion area is artificially covered in the middle of the image (1/3 of the frame), which is used to simulate the situation where the field of view is limited or the camera viewfinder is obscured.
In the sequences, the speaker walks behind the occluded area and then appears on the screen again. For better evaluation, we count the proportion of frames in each sequence where the target was occluded by the mask.

\begin{figure}[t] \small
\centering
\includegraphics[width=1\columnwidth]{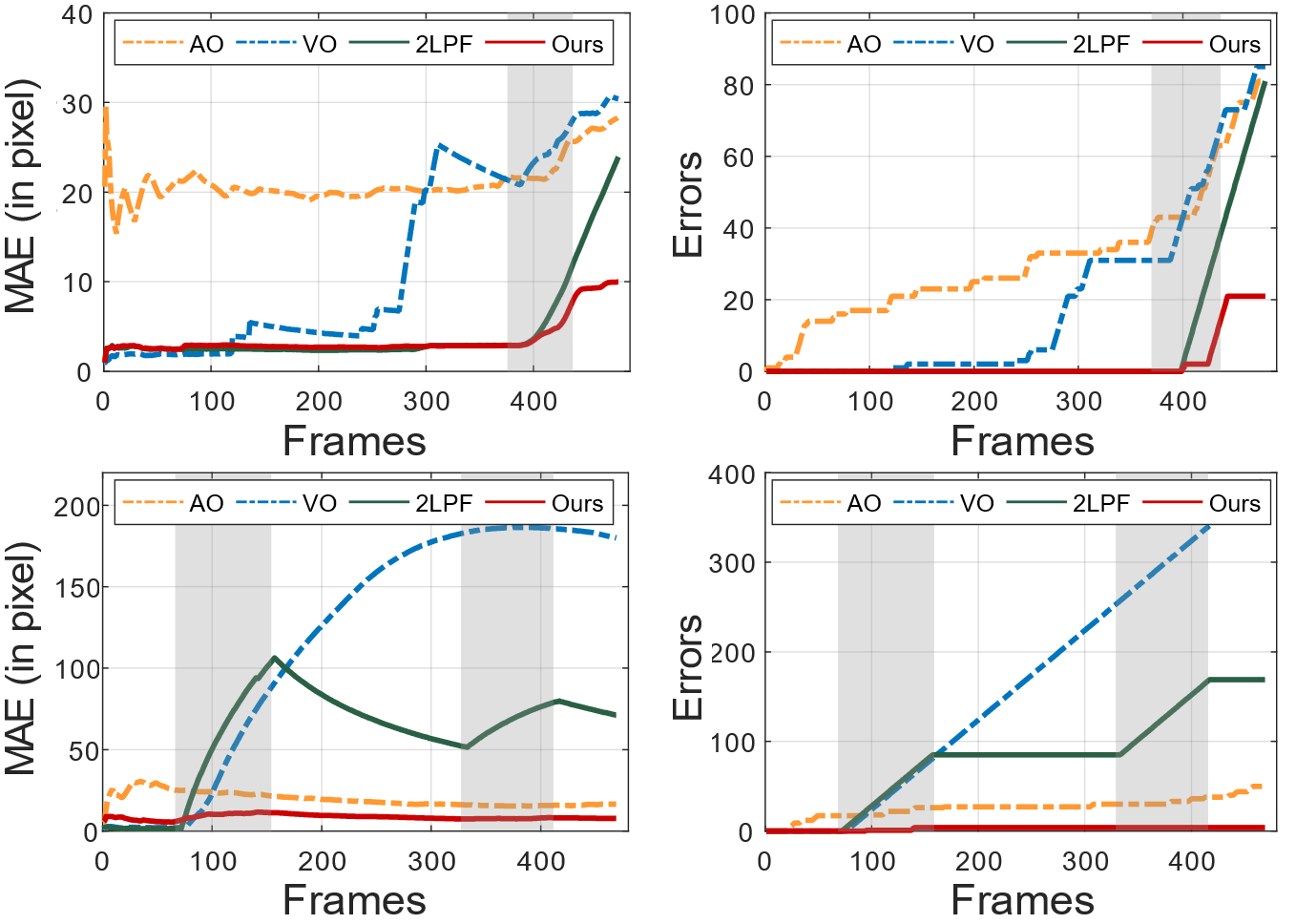} 
\caption{
Tracking accuracy on the Seq11cam1 and Seq08cam2. (Errors: numbers of miss tracking).}
\label{fig:4}
\vspace{-0.4cm}
\end{figure}

Comparison results are listed in Table \ref{table:1}. Firstly, the combination of audio and visual modalities shows great benefits for speaker tracking. 
On the standard dataset, the MAE of the proposed MPT is 4.34 in pixel, which is superior to the state-of-the-art. 2LPF has achieved accurate estimation by employing additional particle filters in audio and visual space, respectively. However, the calculation of fusion likelihood in 2LPF depends on the stable observations, which leads to a rapid decline when visual observation is unavailable.
In contrast, MPT achieves a better tracking accuracy of 78.34\% on sequences with an average occlusion rate of 51.08\%. 
Figure \ref{fig:4} shows the MAE and error numbers of two typical sequences, where the shaded box represents the frames in which the target is occluded. VO and 2LPF are severely affected by occlusion, which can be seen from the significant rise of curves in the shaded area. Our MPT is also affected by occlusion, but the impact is relatively minor.

\begin{figure*}[t]
\centering
\setlength{\abovecaptionskip}{-0.1cm} 
\includegraphics[width=1.9\columnwidth]{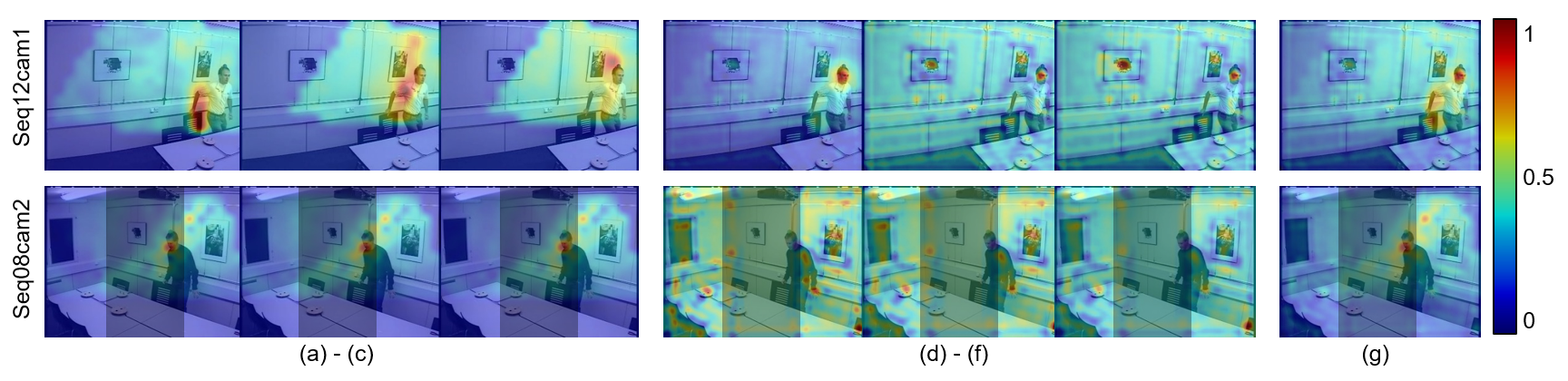} 
\caption{
Visualization of the complementarity of multiple modalities. (a)-(c) are audio cues, (d)-(f) are visual cues, (g) is the fusion map generated according to perception weights. The two rows respectively represent the scene where the audio signal and the video observation are disturbed. The shadow in the middle of the image represents the invisible part.
}
\label{fig:5}
\end{figure*}

\begin{figure*}[t]
\centering
\includegraphics[width=1.9\columnwidth]{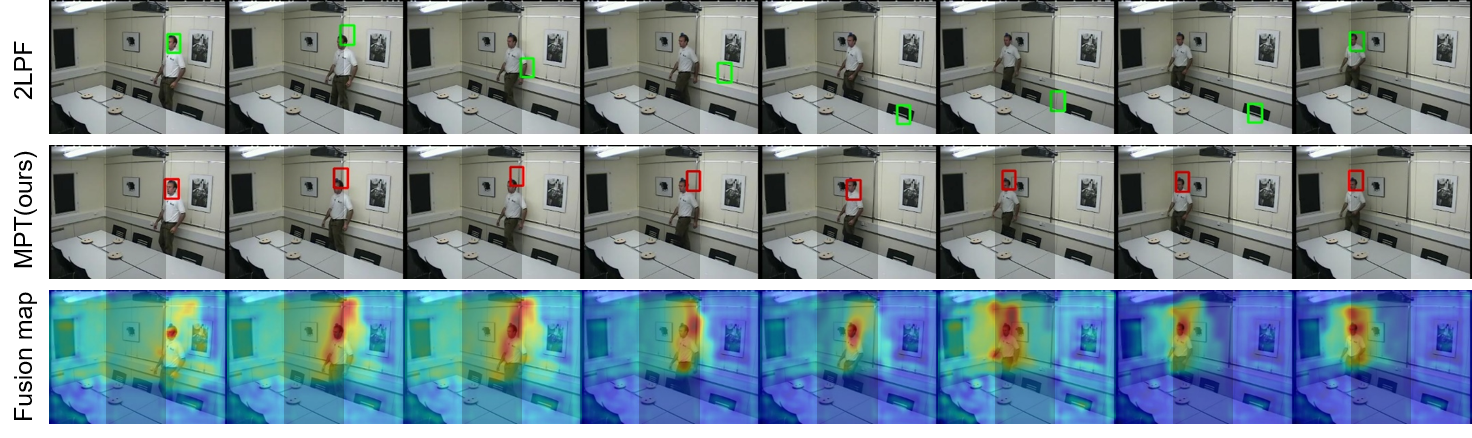} 
\caption{
Comparison of tracking results on the occluded sequence. Green rectangles are from the contrast method 2LPF, and red rectangles come from the proposed MPT method. The bottom row shows the fusion maps of corresponding frames.
}
\label{fig:6}
\end{figure*}

\subsubsection{Ablation Study and Analysis.}
Ablation studies are conducted to evaluate the effectiveness of the main innovative components of our method in Table~\ref{table:2}, including audio measurement, attention network, and PF-based tracker. The general GCF \cite{GCF} calculates the plausibility of the existence of active sound sources at specific coordinates in all possible source positions in a given room. Without the guidance of prior information, it is difficult to derive accurate coordinates within limited calculations. 
In the case of the stGCF method, the search range is reduced to multiple depth planes in 3D space using the projection relationship to achieve sound source localization on the image plane, which has never been studied before.
However, the stGCF method is affected by the geometric configuration of the camera and the microphone array, especially when the speaker is located on the line connecting the camera and the array. 
In addition, due to the directionality of the sound signal, the peak usually appears in a large highlighted area in the stGCF map, which provides an ambiguous search result. 
Nevertheless, the results we calculated using two microphone arrays are better than traditional methods, with the MAE decreasing from 80.15 to 28.40. Note that the result is not changed by visual occlusion. 

In the last four sets of the ablation study, visual cues are added to evaluate the contribution of the attention mechanism. 
The enhancements made by AvgAtt shows the strength of audio-visual fusion, even if it works as a set of weights with the same value. By comparison, our MPAtt achieves higher performance gains. Compared with AvgAtt, the accuracy of MPAtt has increased by 21\% and 19\%, respectively, on the original dataset and the occlusion dataset.
In addition, an improved PF is employed for tracking, which smoothes the trajectory through the time series model. The improved resampling method using the global maximum of the fusion map avoids the particles being restricted to the local optimum due to the target missing of individual frames.

\subsubsection{Visualization Analysis.}
In this section, the audio-visual cues and fusion maps are generated as the heat map to visualize the sub-process of the proposed method, which demonstrates the interpretability of the perception attention network. As illustrated in Figure \ref{fig:5}, in the sample in the first row, the speech is disturbed by the noise emitted by the chair, and in the sample in the second row, the face of the speaker is completely obscured. Nevertheless, the correct area in the fusion map is highlighted. This indicates that benefit from the network's ability to perceive the state of each modality, the model can learn the corresponding perception weights by using the complementarity across the audio-visual cues. Figure \ref{fig:6} shows the robustness of our tracker, which can achieve continuous tracking when the field of view is limited. 
Since the auditory sense is not interfered by the visual distraction, audio cues hold dominance over such difficult samples.
When the speaker walks to the occluded area, the tracker can roughly estimate the speaker's position, which is beneficial to re-track when the target is visible again.

\section{Conclusions}
In this paper, we propose a novel multi-modal perception tracker for the challenging audio-visual speaker tracking task. 
We also propose a new multi-modal perception attention network and a new acoustic map extraction method. 
The proposed tracker utilizes the complementarity and consistency of multiple modalities to learn the availability and reliability of observations between various modalities in a self-supervised fashion.
Extensive experiments demonstrate that the proposed tracker is superior over the current state-of-the-art counterparts, especially showing sufficient robustness under adverse conditions. Lastly, the intermediate process is visualized to demonstrate the interpretability of the proposed tracker network.


\appendix
\label{sec:reference_examples}
\small
\bibliography{aaai22}

\begin{thebibliography}{40}
\providecommand{\natexlab}[1]{#1}

\bibitem[{Afouras et~al.(2021)Afouras, Asano, Fagan, Vedaldi, and
  Metze}]{av-self}
Afouras, T.; Asano, Y.~M.; Fagan, F.; Vedaldi, A.; and Metze, F. 2021.
\newblock Self-supervised object detection from audio-visual correspondence.
\newblock \emph{arXiv:2104.06401}.

\bibitem[{Baltru{\v{s}}aitis, Ahuja, and
  Morency(2018)}]{baltruvsaitis2018multimodal}
Baltru{\v{s}}aitis, T.; Ahuja, C.; and Morency, L.-P. 2018.
\newblock Multimodal machine learning: A survey and taxonomy.
\newblock \emph{IEEE TPAMI}, 41(2): 423--443.

\bibitem[{Ban et~al.(2019)Ban, Alameda-Pineda, Girin, and
  Horaud}]{ban2019variational}
Ban, Y.; Alameda-Pineda, X.; Girin, L.; and Horaud, R. 2019.
\newblock Variational bayesian inference for audio-visual tracking of multiple
  speakers.
\newblock \emph{IEEE TPAMI}, 43(5): 1761--1776.

\bibitem[{Bertinetto et~al.(2016)Bertinetto, Valmadre, Henriques, Vedaldi, and
  Torr}]{simFC}
Bertinetto, L.; Valmadre, J.; Henriques, J.~F.; Vedaldi, A.; and Torr, P.~H.
  2016.
\newblock Fully-convolutional siamese networks for object tracking.
\newblock In \emph{ECCV}, 850--865.

\bibitem[{Brutti and Lanz(2010)}]{brutti2010joint}
Brutti, A.; and Lanz, O. 2010.
\newblock A joint particle filter to track the position and head orientation of
  people using audio visual cues.
\newblock In \emph{European Signal Processing Conference}, 974--978.

\bibitem[{Brutti, Omologo, and Svaizer(2006)}]{GCF}
Brutti, A.; Omologo, M.; and Svaizer, P. 2006.
\newblock Speaker localization based on oriented global coherence field.
\newblock In \emph{ICSLP}, 2606--2609.

\bibitem[{Chiariotti, Martarelli, and
  Castellini(2019)}]{chiariotti2019acoustic}
Chiariotti, P.; Martarelli, M.; and Castellini, P. 2019.
\newblock Acoustic beamforming for noise source localization-reviews,
  methodology and applications.
\newblock \emph{Mechanical Systems and Signal Processing}, 120: 422--448.

\bibitem[{Cobos et~al.(2020)Cobos, Antonacci, Comanducci, and
  Sarti}]{cobos2020frequency}
Cobos, M.; Antonacci, F.; Comanducci, L.; and Sarti, A. 2020.
\newblock Frequency-sliding generalized cross-correlation: A sub-band time
  delay estimation approach.
\newblock \emph{IEEE/ACM TASLP}, 28: 1270--1281.

\bibitem[{Duan et~al.(2021{\natexlab{a}})Duan, Tang, Wang, Zong, Yang, and
  Yan}]{duan2021audio}
Duan, B.; Tang, H.; Wang, W.; Zong, Z.; Yang, G.; and Yan, Y.
  2021{\natexlab{a}}.
\newblock Audio-Visual Event Localization via Recursive Fusion by Joint
  Co-Attention.
\newblock In \emph{WACV}, 4013--4022.

\bibitem[{Duan et~al.(2021{\natexlab{b}})Duan, Wang, Tang, Latapie, and
  Yan}]{duan2021cascade}
Duan, B.; Wang, W.; Tang, H.; Latapie, H.; and Yan, Y. 2021{\natexlab{b}}.
\newblock Cascade attention guided residue learning gan for cross-modal
  translation.
\newblock In \emph{ICPR}, 1336--1343.

\bibitem[{Howard et~al.(2019)Howard, Sandler, Chu, Chen, Chen, Tan, Wang, Zhu,
  Pang, Vasudevan et~al.}]{mobilenetv3}
Howard, A.; Sandler, M.; Chu, G.; Chen, L.-C.; Chen, B.; Tan, M.; Wang, W.;
  Zhu, Y.; Pang, R.; Vasudevan, V.; et~al. 2019.
\newblock Searching for mobilenetv3.
\newblock In \emph{ICCV}, 1314--1324.

\bibitem[{Hu et~al.(2020)Hu, Qian, Jiang, Tan, Wen, Ding, Lin, and
  Dou}]{hu2020discriminative}
Hu, D.; Qian, R.; Jiang, M.; Tan, X.; Wen, S.; Ding, E.; Lin, W.; and Dou, D.
  2020.
\newblock Discriminative sounding objects localization via self-supervised
  audiovisual matching.
\newblock \emph{NeurIPS}, 33.

\bibitem[{Katsaggelos, Bahaadini, and
  Molina(2015)}]{katsaggelos2015audiovisual}
Katsaggelos, A.~K.; Bahaadini, S.; and Molina, R. 2015.
\newblock Audiovisual fusion: Challenges and new approaches.
\newblock \emph{Proceedings of the IEEE}, 103(9): 1635--1653.

\bibitem[{K{\i}l{\i}{\c{c}} et~al.(2015)K{\i}l{\i}{\c{c}}, Barnard, Wang, and
  Kittler}]{kilicc2014audio}
K{\i}l{\i}{\c{c}}, V.; Barnard, M.; Wang, W.; and Kittler, J. 2015.
\newblock Audio assisted robust visual tracking with adaptive particle
  filtering.
\newblock \emph{IEEE TMM}, 17(2): 186--200.

\bibitem[{K{\i}l{\i}{\c{c}} and Wang(2017)}]{kilicc2017audio}
K{\i}l{\i}{\c{c}}, V.; and Wang, W., eds. 2017.
\newblock \emph{Audio-visual speaker tracking}.
\newblock IntechOpen.

\bibitem[{Lathoud, Odobez, and Gatica-Perez(2004)}]{av163}
Lathoud, G.; Odobez, J.-M.; and Gatica-Perez, D. 2004.
\newblock AV16. 3: An audio-visual corpus for speaker localization and
  tracking.
\newblock In \emph{International Workshop on MLMI}, 182--195.

\bibitem[{Li and Qian(2020)}]{av-attention3}
Li, C.; and Qian, Y. 2020.
\newblock Deep audio-visual speech separation with attention mechanism.
\newblock In \emph{ICASSP}, 7314--7318.

\bibitem[{Liu et~al.(2021)Liu, Tang, Latapie, Corso, and Yan}]{liu2021cross}
Liu, G.; Tang, H.; Latapie, H.~M.; Corso, J.~J.; and Yan, Y. 2021.
\newblock Cross-view exocentric to egocentric video synthesis.
\newblock In \emph{ACM MM}, 974--982.

\bibitem[{Liu, Li, and Yang(2019)}]{LYD2LPF}
Liu, H.; Li, Y.; and Yang, B. 2019.
\newblock 3D audio-visual speaker tracking with a two-layer particle filter.
\newblock In \emph{ICIP}, 1955--1959.

\bibitem[{Liu et~al.(2019)Liu, K{\i}l{\i}{\c{c}}, Guan, and
  Wang}]{liu2019audio}
Liu, Y.; K{\i}l{\i}{\c{c}}, V.; Guan, J.; and Wang, W. 2019.
\newblock Audio-visual particle flow smc-phd filtering for multi-speaker
  tracking.
\newblock \emph{IEEE TMM}, 22(4): 934--948.

\bibitem[{Masuyama et~al.(2020)Masuyama, Bando, Yatabe, Sasaki, Onishi, and
  Oikawa}]{masuyama2020self}
Masuyama, Y.; Bando, Y.; Yatabe, K.; Sasaki, Y.; Onishi, M.; and Oikawa, Y.
  2020.
\newblock Self-supervised neural audio-visual sound source localization via
  probabilistic spatial modeling.
\newblock In \emph{IROS}, 4848--4854.

\bibitem[{Omologo and Svaizer(1997)}]{gccphat}
Omologo, M.; and Svaizer, P. 1997.
\newblock Use of the crosspower-spectrum phase in acoustic event location.
\newblock \emph{IEEE TSAP}, 5(3): 288--292.

\bibitem[{Qian et~al.(2019)Qian, Brutti, Lanz, Omologo, and
  Cavallaro}]{qian2019multi}
Qian, X.; Brutti, A.; Lanz, O.; Omologo, M.; and Cavallaro, A. 2019.
\newblock Multi-speaker tracking from an audio-visual sensing device.
\newblock \emph{IEEE TMM}, 21(10): 2576--2588.

\bibitem[{Qian et~al.(2021)Qian, Brutti, Lanz, Omologo, and
  Cavallaro}]{qian2021audio}
Qian, X.; Brutti, A.; Lanz, O.; Omologo, M.; and Cavallaro, A. 2021.
\newblock Audio-visual tracking of concurrent speakers.
\newblock \emph{IEEE TMM}.

\bibitem[{Qian et~al.(2017)Qian, Brutti, Omologo, and Cavallaro}]{qian20173d}
Qian, X.; Brutti, A.; Omologo, M.; and Cavallaro, A. 2017.
\newblock 3D audio-visual speaker tracking with an adaptive particle filter.
\newblock In \emph{ICASSP}, 2896--2900.

\bibitem[{Schymura and Kolossa(2020)}]{schymura2020audiovisual}
Schymura, C.; and Kolossa, D. 2020.
\newblock Audiovisual speaker tracking using nonlinear dynamical systems with
  dynamic stream weights.
\newblock \emph{IEEE/ACM TASLP}, 28: 1065--1078.

\bibitem[{Senocak et~al.(2019)Senocak, Oh, Kim, Yang, and
  Kweon}]{senocak2019learning}
Senocak, A.; Oh, T.-H.; Kim, J.; Yang, M.-H.; and Kweon, I.~S. 2019.
\newblock Learning to localize sound sources in visual scenes: Analysis and
  applications.
\newblock \emph{IEEE TPAMI}.

\bibitem[{Tang et~al.(2021)Tang, Liu, Xu, Torr, and
  Sebe}]{tang2021attentiongan}
Tang, H.; Liu, H.; Xu, D.; Torr, P.~H.; and Sebe, N. 2021.
\newblock Attentiongan: Unpaired image-to-image translation using
  attention-guided generative adversarial networks.
\newblock \emph{IEEE TNNLS}.

\bibitem[{Tang et~al.(2019)Tang, Xu, Sebe, Wang, Corso, and
  Yan}]{tang2019multi}
Tang, H.; Xu, D.; Sebe, N.; Wang, Y.; Corso, J.~J.; and Yan, Y. 2019.
\newblock Multi-channel attention selection gan with cascaded semantic guidance
  for cross-view image translation.
\newblock In \emph{CVPR}, 2417--2426.

\bibitem[{Wang et~al.(2018)Wang, Teng, Xing, Gao, Hu, and
  Maybank}]{attention2018}
Wang, Q.; Teng, Z.; Xing, J.; Gao, J.; Hu, W.; and Maybank, S. 2018.
\newblock Learning attentions: Residual attentional siamese network for high
  performance online visual tracking.
\newblock In \emph{CVPR}, 4854--4863.

\bibitem[{Wilson and Lin(2020)}]{wilson2020avot}
Wilson, J.; and Lin, M.~C. 2020.
\newblock AVOT: Audio-visual object tracking of multiple objects for robotics.
\newblock In \emph{ICRA}, 10045--10051.

\bibitem[{Woo et~al.(2018)Woo, Park, Lee, and Kweon}]{CBAM}
Woo, S.; Park, J.; Lee, J.-Y.; and Kweon, I.~S. 2018.
\newblock Cbam: Convolutional block attention module.
\newblock In \emph{ECCV}, 3--19.

\bibitem[{Xu et~al.(2018)Xu, Wang, Tang, Liu, Sebe, and
  Ricci}]{xu2018structured}
Xu, D.; Wang, W.; Tang, H.; Liu, H.; Sebe, N.; and Ricci, E. 2018.
\newblock Structured attention guided convolutional neural fields for monocular
  depth estimation.
\newblock In \emph{CVPR}, 3917--3925.

\bibitem[{Xuan et~al.(2020)Xuan, Zhang, Chen, Yang, and Yan}]{xuan2020cross}
Xuan, H.; Zhang, Z.; Chen, S.; Yang, J.; and Yan, Y. 2020.
\newblock Cross-modal attention network for temporal inconsistent audio-visual
  event localization.
\newblock In \emph{AAAI}, 279--286.

\bibitem[{Yang et~al.(2019)Yang, Liu, Pang, and Li}]{yang2019multiple}
Yang, B.; Liu, H.; Pang, C.; and Li, X. 2019.
\newblock Multiple sound source counting and localization based on tf-wise
  spatial spectrum clustering.
\newblock \emph{IEEE/ACM TASLP}, 27(8): 1241--1255.

\bibitem[{Yang et~al.(2021)Yang, Tang, Ding, Sebe, and
  Ricci}]{yang2021transformer}
Yang, G.; Tang, H.; Ding, M.; Sebe, N.; and Ricci, E. 2021.
\newblock Transformer-based attention networks for continuous pixel-wise
  prediction.
\newblock In \emph{ICCV}, 16269--16279.

\bibitem[{Yu et~al.(2020)Yu, Xiong, Huang, and Scott}]{attention2020}
Yu, Y.; Xiong, Y.; Huang, W.; and Scott, M.~R. 2020.
\newblock Deformable siamese attention networks for visual object tracking.
\newblock In \emph{CVPR}, 6728--6737.

\bibitem[{Zeng, Wang, and Lu(2019)}]{attention2019}
Zeng, Y.; Wang, H.; and Lu, T. 2019.
\newblock Learning spatial-channel attention for visual tracking.
\newblock In \emph{ICCC}, 277--282.

\bibitem[{Zhang et~al.(2016)Zhang, Chen, Rasch, and
  Wu}]{zhang2016decentralized}
Zhang, W.-H.; Chen, A.; Rasch, M.~J.; and Wu, S. 2016.
\newblock Decentralized multisensory information integration in neural systems.
\newblock \emph{Journal of Neuroscience}, 36(2): 532--547.

\bibitem[{Zhao et~al.(2020)Zhao, Ma, Gu, Yang, Xing, Xu, Hu, Chai, and
  Keutzer}]{av-attention2}
Zhao, S.; Ma, Y.; Gu, Y.; Yang, J.; Xing, T.; Xu, P.; Hu, R.; Chai, H.; and
  Keutzer, K. 2020.
\newblock An end-to-end visual-audio attention network for emotion recognition
  in user-generated videos.
\newblock In \emph{AAAI}, 303--311.

\end{thebibliography}
\end{document}